\documentclass[letterpaper, 10 pt, conference]{ieeeconf}  
\IEEEoverridecommandlockouts                              
\usepackage{bm}
\usepackage{times}
\usepackage{epsfig}
\usepackage{amssymb}
\usepackage{algpseudocode}
\usepackage{algorithm}
\usepackage{amsmath}
\usepackage{subfigure}
\usepackage{graphicx}
\usepackage{multirow}
\usepackage{multicol}
\usepackage{caption}
\usepackage[pdfstartview=FitH,bookmarksnumbered=true,colorlinks,bookmarksopen=true]{hyperref}
\usepackage{mathtools}
\usepackage{booktabs}
\usepackage{pifont}

\IEEEoverridecommandlockouts                              
\usepackage{array,multirow,graphicx} 
\usepackage{epsfig} 
\usepackage{mathptmx} 
\usepackage{times} 
\usepackage{amsmath} 
\usepackage{amssymb}  
\usepackage{gensymb}
\usepackage{float}
\usepackage{xcolor}
\usepackage[normalem]{ulem}
\usepackage{wrapfig}
\usepackage{booktabs}
\usepackage{stfloats}
\title{\LARGE \bf
Learning Tri-mode Grasping for Ambidextrous Robot Picking
}

\author{Chenlin Zhou\textsuperscript{1, 2}, Peng Wang\textsuperscript{1, 2, 3\ding{41}}, Wei Wei\textsuperscript{1, 2},   Guangyun Xu\textsuperscript{1,2}, Fuyu Li\textsuperscript{1,2}, Jia Sun\textsuperscript{1} 
\thanks{This work was supported in part by the National Natural Science Foundation of China under Grants (91748131, 62006229 and 61771471), in part by the Strategic Priority Research Program of Chinese Academy of Science under Grant XDB32050100, and in part by the InnoHK Project.}
\thanks{$^{1}$ Institute of Automation, Chinese Academy of Sciences, Beijing, China.} 
\thanks{$^{2}$ School of Artificial Intelligence, University of Chinese Academy of Sciences, Beijing, China}
\thanks{$^{3}$ Centre for Artificial Intelligence and Robotics, Hong
Kong Institute of Science and Innovation, Chinese Academy of Sciences, Hong Kong, China}
\thanks{\ding{41} Corresponding author:  peng\_wang@ia.ac.cn}
}

\begin{document}
\maketitle
\thispagestyle{empty}
\pagestyle{empty}

\begin{abstract}
Object picking in cluttered scenes is a widely investigated field of robot manipulation, however, ambidextrous robot picking is still an important and challenging issue. We found the fusion of different prehensile actions (grasp and suction) can expand the range of objects that can be picked by robot, and the fusion of prehensile action and nonprehensile action (push) can expand the picking space of  ambidextrous robot. In this paper, we propose a Push-Grasp-Suction (PGS) tri-mode grasping learning network for ambidextrous robot picking through the fusion of different prehensile actions and the fusion of prehensile action and nonprehensile action.
The prehensile branch of PGS takes point clouds as input, and the 6-DoF picking configuration of grasp and suction in cluttered scenes are generated by multi-task point cloud learning. The nonprehensile branch with depth image input generates instance segmentation map and push configuration, cooperating with the prehensile actions to complete the picking of objects out of single-arm space. PGS generalizes well in real scene and achieves state-of-the-art picking performance. 

\end{abstract}

\section{INTRODUCTION}
Robot picking is a fundamental but challenging topic for intelligent robotic manipulation, and has a great application in the fields of manufacturing \cite{pedersen2016robot},
medical services \cite{kumar2018job, garcia2019variability,nakamura2019ontology} and so on. 
However, there are still many challenging and important problems in unstructured picking applications, such as limited range of pickable objects, poor generalization of picking algorithms, limited working space of the dual-arm robot and so on, which bring great challenge for the dexterous picking ability of existed picking methods.

\begin{figure}[tbp]
    \centering
    \includegraphics[width=1.\linewidth]{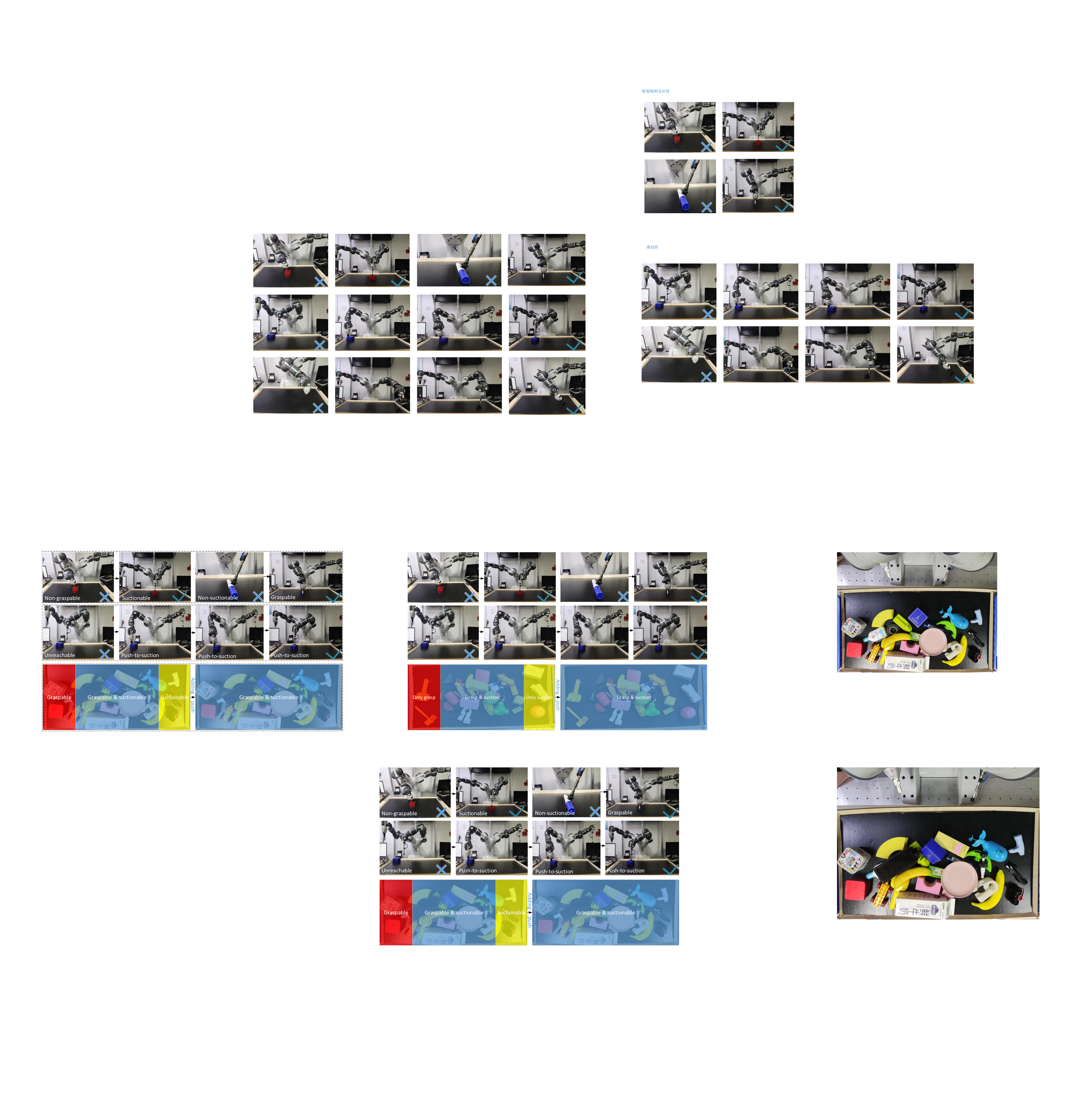}
    \vspace{+0.1in}
    \caption{Illustrative configuration of prehensile actions and nonprehensile action. First line shows the complementarity of different prehensile actions (grasp and suction). Second line shows nonprehensile action (push) assisted suction for the  picking of object out of suction-arm space. Last line shows combining prehensile actions and nonprehensile action could expand the picking space for ambidextrous robot.
    }
    \label{fig:idea}
    \vspace{-6mm}
\end{figure}

Combining different prehensile actions (grasp and suction) is an effective way to expand the range of objects that can be picked by robot, improving the adaptability of robot system \cite{hasegawa2017three, nakamoto2018gripper, deng2019deep}. 
As illustrated in the first line of Fig.\ref{fig:idea},
Parallel-jaw gripper based grasp and silicone suction-cup based suction complement each other, 
the former can easily pick small objects, such as paper clips, watercolor pens \cite{stuart2018tunable, xu2021pois}, while the latter can easily pick larger objects with planar surfaces such as boxes. Therefore, combining them increases robotic adaptability for objects with various geometries and materials, expanding the range of pickable objects.
In addition, for the dual-arm robot, there is a problem that objects is unreachable for the picking action (grasp or suction ) in the picking space of dual-arm robot. As illustrated in the second line of Fig.\ref{fig:idea}. Suction action can not reach the object, a nonprehensile action (push action) is introduced to assist suction action to complete the picking of object. Therefore, push action could expand the picking space of ambidextrous robot through push-to-suction or push-to-grasp way. 

Based on the above analysis, in this paper, we proposes a Push-Grasp-Suction (PGS) tri-mode grasping learning network for ambidextrous robot picking through the fusion of different prehensile actions and the fusion of prehensile action and nonprehensile action.
PGS takes point cloud or depth image as input to ensure the picking generalization, without RGB information.
The prehensile branch of PGS takes point clouds as input, and the 6-DoF picking configuration of grasp and suction in cluttered scenes are generated by multi-task point cloud learning. The nonprehensile branch with depth image input generates instance segmentation map and push configuration, cooperating with the prehensile actions to complete the picking of objects out of single-arm space. Our main contributions are as follows:
 
1. A push-grasp-suction tri-mode grasping learning network for ambidextrous robot picking  is established through the fusion of different prehensile actions and the fusion of prehensile action and nonprehensile action. 

2. With the proposed push-grasp-suction tri-mode learning network, robot system could carry out many ways (grasp, suction, push-to-grasp, push-to-suction) to clear objects in cluttered scene effectively.

3. Our proposed method outperforms state-of-the-art robot picking methods both on large-scale dataset and real robot experiments.

\section{RELATED WORK}
In this section, we introduce the previous works related to ours.

\textbf{Deep  Learning  based  Picking  Configuration  Detection.}
Model-free 6-DoF picking (grasp or suction) has become a research hot spot and dominant direction in robot picking. This way generates the 6-DoF picking configuration of object directly from the sensor output data and do not require  prior knowledge about the geometry information of object. 
Model-free 6-DoF picking method has better generalization and dexterity, compared with traditional methods (Model-based 6-DoF picking\cite{xiang2017posecnn,wang2019densefusion,deng2020self}, Top-down picking\cite{depierre2018jacquard,lenz2014deep,guo2016object,jiang2011efficient,redmon2015real, pinto2016supersizing,mahler2017dex} ).

GPD \cite{ten2017grasp} can generate grasp poses on any visible surface, relative to prior methods does not require a precise segmentation of the object to be grasped.
PointNetGPD \cite{liang2019pointnetgpd} is an end-to-end  lightweight grasp evaluation model which  can capture the complex geometric structure of the contact area between gripper and object. It directly process the 3D point cloud that locates within the gripper for grasp evaluation.
GPR \cite{wei2021gpr} is a two-stage grasp pose refinement network which detects grasps globally while fine-tuning low-quality grasps and filtering noisy grasps locally. In addition, it  also extend the 6-DoF grasp with an extra dimension as grasp width which is critical for collisionless grasping in cluttered scenes.
\cite{li2021simultaneous} formalizes the 6-DoF grasp pose estimation as a simultaneous multi-task learning problem, jointly predicts the feasible 6-DoF grasp poses, instance semantic segmentation, and collision information.
SuctionNet \cite{cao2021suctionnet} is a new physical model to analytically evaluate seal formation and wrench resistance of a suction pose, which are two key aspects of suction success. These methods belong to a single-type prehensile action method, which has limited range of pickable objects.

Above methods are all focused on single-prehensile action, however, research on muti-type prehensile action method is very rare. Different prehensile actions (such as grasp and suction) fusion can expand the range of objects that can be picked by robot, improving the success rate and completion rate of muti-objects picking scene. It is really worthy of further exploration. For muti-type (two-type) prehensile action method,
POIS\cite{xu2021pois} takes depth image as input and generate a pair of target masks that allows ambidextrous robots to pick in parallel through other picking methods.
DexNet 4.0 \cite{mahler2019learning} can be trained on synthetic data  and uses Grasp Quality Convolutional Neural Network (GQ-CNN) to predict the grasp qualities of parallel-jaws and suction cups separately, then select a picking way by a RL model to maximize pciking quality. Both POIS and DexNet 4.0 are all muti-step methods, can not generate the 6-DoF picking configuration of object from the sensor output data directly. Our proposed PGS formalize the 6-DoF grasp and suction pose estimation as a simultaneous multi-task learning problem, is an end-to-end way to generate muti-type prehensile actions' configration.
 
\textbf{Picking Dataset Generation.}
GPR \cite{wei2021gpr} build a synthetic single-object grasp dataset including 150 commodities of various shapes, and a complex multi-object cluttered scene dataset including 100k point clouds with robust, dense grasp poses and mask annotations.
POIS \cite{xu2021pois}  builds a synthetic dataset includes RGB, depths, instance segmentation masks for robot picking (grasp and suction).
To deal with the gap between the virtual environment and the real world, \cite{fang2020graspnet} constructs a general grasp dataset in cluttered scenes, in which images are captured by regular commodity cameras. 

\textbf{Deep Learning for Point Cloud.} 
PointNet \cite{qi2017pointnet}  and PointNet++ \cite{qi2017PointNet++} are two novel network to directly extract feature representation from point cloud data. Such a point-based network has been widely used in the 3D domain such as classification, segmentation and detection \cite{tatarchenko2017octree,graham20183d,shi2019pointrcnn}. The prehensile branches of our PGS  adopt PointNet++ as our backbone to extract point-wise features from disorderly point sets.

\textbf{Grasping  Assisted by Push Action.}
Combining nonprehensile and prehensile action aimed at rearanging  objects by pushing to simplify the scene, so that prehensile action could be carried out to pick object . 
VPG \cite{zeng2018learning} propose to discover and learn synergies
between pushing and grasping from experience through model-free deep reinforcement learning.
\cite{deng2019deep} utilize the action  pushing motions to actively explore and change the environment on the basis of the affordance map. 
Current grasping method assisted by push action are all belongs to single-arm robot with single-type prehensile action method.
In addition, these methods aim at picking adhesion and obscure objects, scenes are with small number objects and few types of objects.
The picking part of these methods are mainly Top-down way, can not meet the dexterity for picking complex and diverse objects in clutter.
Our proposed PGS are the first to combine the fusion of different
prehensile actions and the fusion of prehensile action and
nonprehensile action in object picking, aimed at improve robotic dexterous picking ability in the scene with cluttered and diverse objects.


\section{METHOD}
\begin{figure*}[tp]
	\centering
	\includegraphics[width=1.\textwidth]{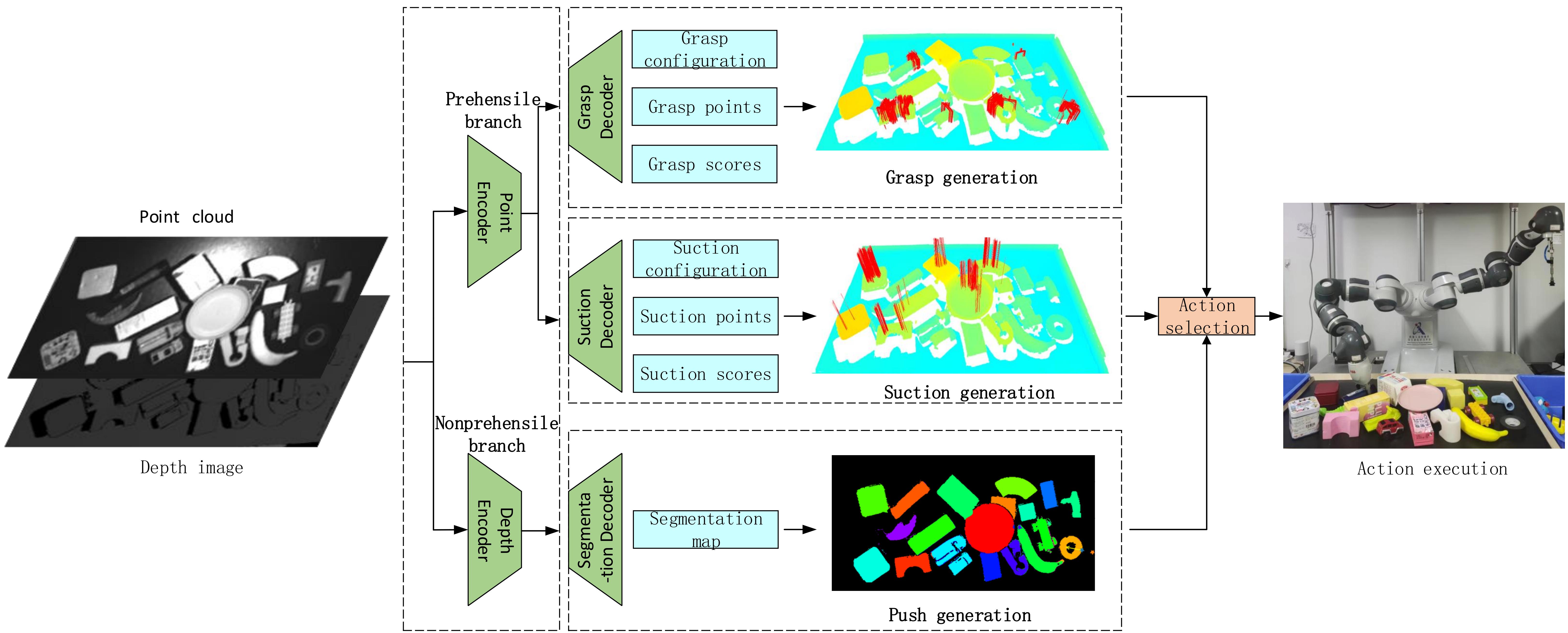}
	\vspace{+0.2cm}
	\caption{The pipeline of push-grasp-suction tri-mode learning network for ambidextrous robot picking. On prehensile action branches, given a scene point cloud which can be converted by a scene depth image, we use a point encoder to extract global features of scene, then, simultaneously attach two parallel point decoders for 6-DoF grasp pose generation and  6-DoF suction pose generation. On nonprehensile action branch,  given a scene depth image, we used an improved U-net architecture to generate the segmentation map which can be utilized the top-down push pose. Robot system could carry out many ways (grasp, suction, push-to-grasp, push-to-suction) to clear objects in cluttered scene effectively.}
	\label{fig:method}
	\vspace{-0.6cm}
\end{figure*}

Our proposed PGS contains prehensile branch (6-DoF grasp generation and 6-DoF suction generation) and nonprehensile branch (instance segmentation and top-down push generation). Given a scene point cloud which can be converted by a scene depth image, we use a backbone network PointNet++\cite{qi2017PointNet++} to encode global features of scene, Then, simultaneously attach two parallel decoders for 6-DoF grasp pose and  6-DoF suction pose. On nonprehensile action branch,  given a scene depth image, we used an improved U-net architecture to generate the segmentation map which can be utilized the top-down push pose. In this section, the details of PGS are discussed. The schematic diagram of PGS is shown in Fig.\ref{fig:method}.

\subsection{Grasp Generation}
Similar to we previous work GPR\cite{wei2021gpr}, our grasp generation branch predicts 6-DoF grasp pose + grasp width for each point directly in the scene point clouds. We utilize the PointNet++ network with multi-scale grouping strategy as the backbone, which is a robust learning model for dealing with sparse point cloud and non-uniform point density. 

\begin{figure}
    \centering
    \subfigure[]{\includegraphics[width=0.4\linewidth]{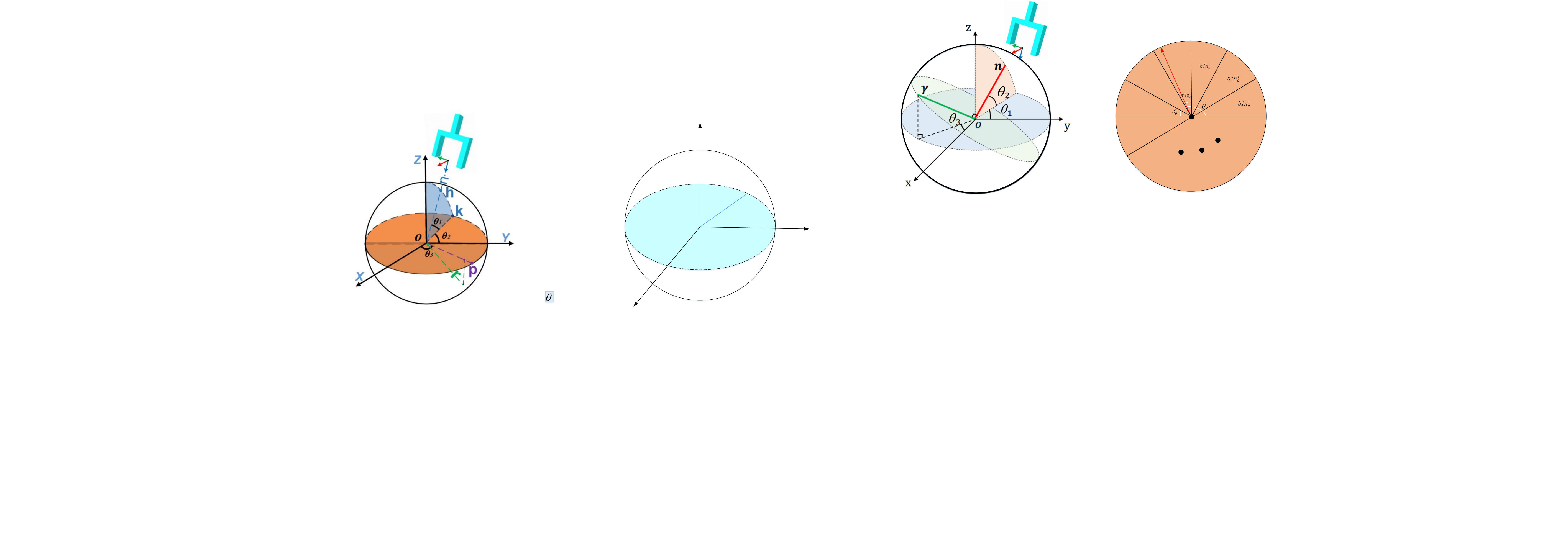}
    \label{fig:bin_base_grasp_a}}
    \subfigure[]{\includegraphics[width=0.45\linewidth]{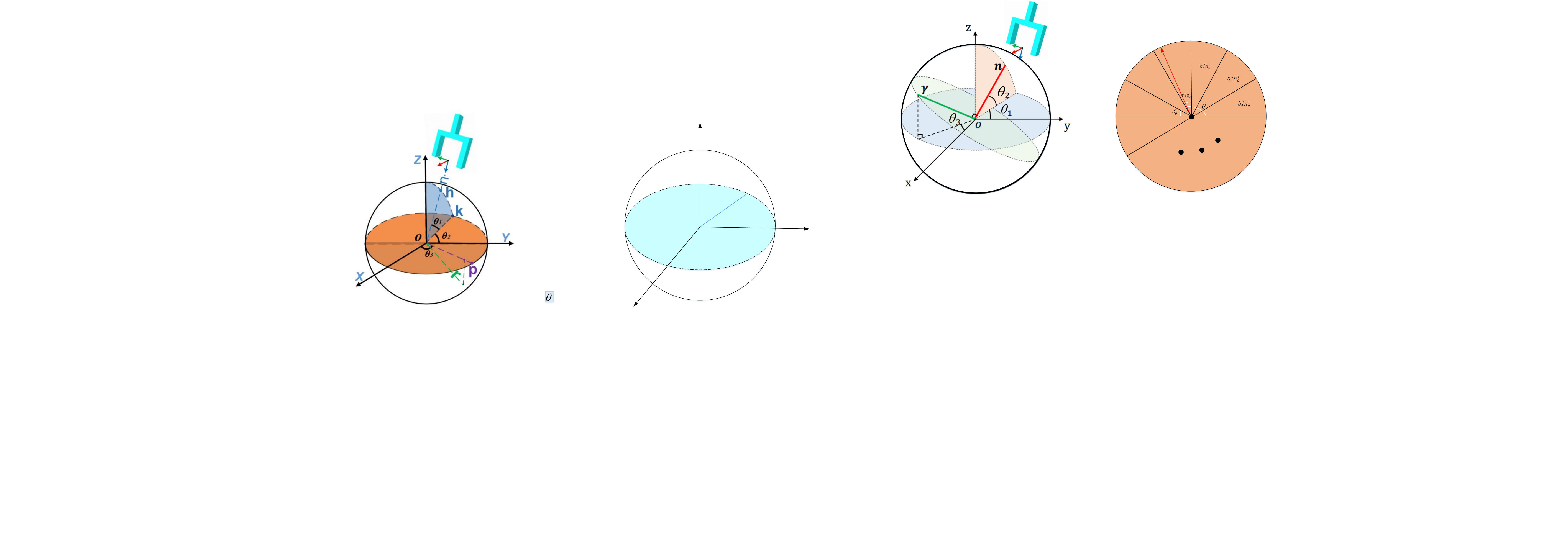}
    \label{fig:bin_base_grasp_b}}
    \caption{An illustration of bin-based regression and picking rotation.
    \subref{fig:bin_base_grasp_a} shows $\theta$ is encodes as bin classification target $\operatorname{bin}_{\theta}$ and residual regression target $\operatorname{res}_{\theta}$ in bin-based regression. 
    \subref{fig:bin_base_grasp_b} The grasp approach vector $\textbf{n}$ is denoted by azimuth angle \textbf{$\theta_1$} and elevation angle \textbf{$\theta_2$}, 
    grasp  axis \textbf{r} is  located  on  the  plane orthogonal to the approach vector  and its projection on X-Y plane is \textbf{$\theta_3$}. 
   }
    \label{fig:bin_base_grasp}
    \vspace{-4mm}
\end{figure}

\textbf{Bin-based Grasp Configuration Regression.} It is difficult to regress 6-DoF grasp pose directly, which has been proved in previous literature\cite{kehl2017ssd,qi2019deep, shi2019pointrcnn}. Therefore, we utilize a bin-based regression method for 6-DoF grasp pose and grasp width  regression\cite{wei2021gpr, zhou2021bv,zhou2021acr}, which  transforms a parameter regression problem into N regression N classification. 
A target regression parameter $\bm o$ can be encoded as bin classification target $\operatorname{bin}_{\bm o}:\left[\operatorname{bin}_{\bm o}^{1}, \operatorname{bin}_{\bm o}^{2}, \ldots, \operatorname{bin}_{\bm o}^{m}\right]$ and   residual regression target
$\operatorname{res}_{\bm o}:\left[\operatorname{res}_{\bm o}^{1}, \operatorname{res}_{\bm o}^{2}, \ldots, \operatorname{res}_{\bm o}^{m}\right]$.
The schematic diagram is shown in Fig.\ref{fig:bin_base_grasp}\subref{fig:bin_base_grasp_a}, which shows how to regress a angle parameter $\theta \in[0,2 \pi]$. $\delta_{\theta}$ is the unit bin angle of $\theta$ for normalization. The  bin classification target $\operatorname{bin}_{\theta}=\left\lfloor\theta / \delta_{\theta}\right\rfloor$, the  residual regression target
$\operatorname{res}_{\theta}=\left(\theta-b i n_{\theta} * \delta_{\theta}\right) / \delta_{\theta}$.
In the example of Fig.\ref{fig:bin_base_grasp}\subref{fig:bin_base_grasp_a}, the $\operatorname{bin}_{\theta}$ and $\operatorname{res}_{\theta}$ can be encoded as $[0,0,1, \ldots, 0]$ and $\left[0,0, \operatorname{res}_{\theta}, \ldots, 0\right]$.
In addition, the network predicted parameter are also encode as bin classification part  and  residual regression part. Then, the cross entropy loss and MSE loss can be carried out between the predicted parameter and the ground-truth.

A grasp configuration (6-DoF grasp pose and grasp width) is represented as $\bm g =  (\bm L,  \bm \Theta, w)$, where $\bm L = (x, y, z)$ denotes the grasp center, $\bm \Theta = \theta_{1}, \theta_{2}, \theta_{3}$ denotes the rotation representation of gripper. $w$ denotes the opening width of gripper. 
We define $\bm \Theta = \theta_{1}, \theta_{2}, \theta_{3}$ through grasp approach vector $\bm n$ and grasp axis $\bm r$, which is shown in Fig.\ref{fig:bin_base_grasp}\subref{fig:bin_base_grasp_b}. Azimuth angle $\theta_{1} \in [0,2 \pi]$ and elevation angle $\theta_{2} \in [0, \pi / 2]$ jointly define the grasp approach vector $\bm n$. Rotation angle $\theta_{3} \in [- \pi, \pi]$ denotes the projected axis  r onto X-Y plane. 
All the grasp configuration parameters are regress by bin-based regression method. The  encode process of $\theta_{1}, \theta_{2}, \theta_{3}$ can be formulated as follows:

\begin{equation}
\operatorname{bin}_{\theta}^{p}=\left\lfloor\frac{\theta^{p}-\theta_{s}}{\delta_{\theta}}\right\rfloor, \theta \in\left\{\theta_{1,2,3}\right\}\label{bin_encode}
\end{equation}
\begin{equation}
r e s_{\theta}^{p}=\frac{1}{\delta_{\theta}}\left(\theta^{p}-\theta_{s}-\left(\operatorname{bin}_{\theta}^{p} \times \delta_{\theta}\right)\right), \theta \in\left\{\theta_{1,2,3}\right\}\label{res_encode}
\end{equation}

Where $\delta_{\theta}$ is the uniform angle, $\theta_{s}$ denote the starting angle. For grasp center and grasp width prediction, we adopt the following formulation:
\begin{equation}
\operatorname{bin}_{u}^{p}=\left\lfloor\frac{u^{p}-u^{(p)}}{d_{u}}\right\rfloor, u \in\{x, y, z, w\}
\end{equation}
\begin{equation}
r e s_{u}^{p}=\frac{1}{d_{u}}\left(u^{p}-u^{(p)}-\left(\operatorname{bin}_{u}^{p} \times d_{u}\right)\right), u \in\{x, y, z, w\}
\end{equation}

Where $d_{u}$ is the uniform bin length, $x_{u},y_{u},z_{u}$ is the coordinates of a interest grasp contact point. $w_{u}$ is the start grasp width. The overall loss of grasp configuration generation could be formulated as follows:
\begin{equation}
\begin{aligned}
loss _{g r a s p}^{p}=\frac{1}{N_{p o s}} \sum_{p \in p o s} \sum_{u \in\left\{x, y, z, w, \theta_{1,2,3}\right\}}(l_{c l s}(\widehat{\operatorname{bin}}_{u}^{p}, \operatorname{bin}_{u}^{p}) \\
 +l_{r e g} (\widehat{\operatorname{res}}_{u}^{p}, r e s_{u}^{p}) )\label{grasp_loss}
\end{aligned}
\end{equation}

Where $\widehat{\operatorname{bin}}_{u}^{p}$ and $\widehat{\operatorname{res}}_{u}^{p}$ are the predicted bin assignment and residual assignment. $bin_{u}^{p}$ and $r e s_{u}^{p}$ are the corresponding ground-truth. $l_{c l s}$ and $l_{r e g}$ represent cross entropy loss and Mean Square Error (MSE) loss, respectively.

In addition, we use focal loss $loss _{ grasp }^{focal}$ \cite{lin2017focal} for handle the severe imbalance problem between predicted grasp points and negative points. The predicted grasp scores use MSE loss, recorded as $loss _{ grasp }^{score}$. Note that the ground-truth of grasp and suction score are described in \ref{dataset}. Therefore, the total loss $loss _{ grasp}$ in grasp action could be formulated as follows:
\begin{equation}
loss _{ grasp} = loss _{ grasp }^{p} +  loss _{ grasp }^{focal} +  loss _{ grasp }^{score}\label{total_grasp_loss}
\end{equation}

\subsection{Suction Generation}

Suction generation part shares PointNet++ backbone with grasp generation part for point feature extraction.
6-DoF suction configuration  also  adopt  bin-based regression method.

\textbf{Bin-based Suction configuration Regression.} 
A suction configuration do not have the width of gripper, so  a 6-DoF suction configuration is represented as $\bm g =  (\bm L,  \bm \Theta)$, where $\bm L = (x, y, z)$ denotes suction center. $\bm \Theta = \theta_{1}, \theta_{2}, \theta_{3}$ denotes the normal direction of suction.
In suction configuration, suction center coincides with suction contact point, which  is different from grasp configuration. 
Our network is point-wise proposal for suction configuration, therefore, suction pose configuration only regress the normal direction $\theta_{1}, \theta_{2}, \theta_{3}$ of the candidate suction contact point. 
The  encode process of $\theta_{1}, \theta_{2}, \theta_{3}$ are same with rotation encoding of grasp pose, which is shown in Eq.\ref{bin_encode} and Eq.\ref{res_encode}. Then, we could easily calculate the loss of suction proposal generation $loss_{ suction }^{p}$ through Eq.\ref{grasp_loss}. Note that $u \in\{\theta_{1,2,3}\}$ when calculate $loss_{ suction }^{p}$. 

The total loss $loss _{suction}$ in suction action could be calculated like Eq.\ref{total_grasp_loss}. Note that $loss _{ suction }^{focal}$ and $  loss _{ suction }^{score}$ are same with grasp action.






\textbf{Prehensile branch loss.} 
The prehensile branch is essentially a multitask learning network for grasp and suction action. So, the total loss for prehensile branch of PGS is shown as follow:
\begin{equation}
loss _{ pre} = loss _{ grasp } + loss_{ suction }
\end{equation}

\subsection{Instance Segmentation and Push Generation}
The nonprehensile branch with depth image input generates instance segmentation map and
push configuration, cooperating with the prehensile actions to complete the picking of objects out of single-arm space. Instance segmentation map could count object number and picking progress of scene in real time, improving the system interactivity. In addition, it can be utilized the top-down push configuration.

\textbf{Network Structure.} 
The backbone is an improved U-Net \cite{ronneberger2015u}, which is an encoder-decoder network structure for 3D reason in UOIS \cite{xie2021unseen}.
In preprocessing, the depth image is converted into 3-channel organized point cloud, $ D \in \mathbb{R}^{H \times W \times 3}$, of $XYZ$ coordinates by backprojecting a depth image with camera intrinsics. In backbone structure, we use a 
the improved U-Net added a GroupNorm layer and Relu after each $3 *3$ convolutional layer. In addition, the improved U-Net use dilated convolutions \cite{yu2015multi} for providing model with a higher receptive field. To making model computationally efficient, the improved U-Net also replace the $6_{th},$ $8_{th},$ and $10_{th}$convolution layers with ESP modules \cite{mehta2018espnet}, which has less parameters than the convolution layer.
The network outputs 3D offsets to object centers $V^{\prime} \in \mathbb{R}^{H \times W \times 3}$. Therefore, $D + V^{\prime}$ is the predicted object centers for each pixel. 

\textbf{Loss Function.} The loss function of network consists of foreground loss and  center offset loss. We use a weighted cross entropy  in classifying background and foreground in depth image. The mathematical expression of $l_{f g}$  is as follows:
\begin{equation}
l_{f g}=\sum_{i \in \Omega} w_{i} l_{c e}\left(\hat{F}_{i}, F_{i}\right)
\end{equation}
where $i$ ranges over pixels, $\hat{F}_{i}$, $F_{i}$ are the predicted and ground truth probabilities of pixel i, respectively. $l_{c e}$ is the cross entropy loss. The weight $w_{i}$ is inversely proportional to the number of pixels with labels equal to $\hat{F}_{i}$, normalized to sum to 1.
We apply a Huber loss $\rho$ (Smooth L1 loss) to the center offsets $V_{i}^{\prime}$ to penalize the distance of the center votes to their corresponding ground truth  object centers. The mathematical expression of $l_{f g}$  is as follows:
\begin{equation}
l_{c o}=\sum_{i \in \Omega} w_{i} \rho\left(D_{i}+V_{i}^{\prime}-c_{i}\right)
\end{equation}
where $c_{i}$ is the 3D coordinate of the ground truth object center for pixel i. The weight $w_{i}$ is inversely proportional to the number of pixels with the same instance label.
In summary, the total loss of nonprehensile branch is given by:
\begin{equation}
loss_{nonpre} =\lambda_{f g} l_{f g}+\lambda_{c o} l_{c o}
\end{equation}
Where $\lambda_{f g}$,$\lambda_{c o}$ represent the balance weights for different losses.
After network center votes learning, a meanshift \cite{kong2018recurrent} clustering algorithm will be applied to group points belong to same instance.

\textbf{Push Generation.} Push action can be generate by segmentation results and rule-based method. Push configuration in image can be represented as $(x, y, \theta, d)$, $(x, y)$ represents push starting coordinates, $\theta$ represents push direction which can be define by target object towards scene center, $d$ represents push distance. When executing, it can be translate $(x_{1}, y_{1}, x_{2}, y_{2})$, represents push starting coordinates and push ending coordinates. 
We choose the farthest point $(x_{0}, y_{0})$  of object which pre-execution prehensile action can not reach, along push direction. To avoid end effector colliding with object, the $(x_{0}, y_{0})$ need adding a protecting distance to get push starting coordinates $(x, y)$.

\subsection{Action Selection}
We rank the output prehensile action (grasps and suctions)  by prehensile action scores which the network predicts. Then, the max action score of prehensile action could be choose to execute. When prehensile action cannot reach target location, nonprehensile action (push) could be executed to move object towards scene center until prehensile action can be executed correctly.

\section{Dataset}\label{dataset}
We extended the grasp dataset of GPR \cite{wei2021gpr} through adding suction annotation in cluttered scenes, so our dataset includes 100k point clouds of multi-object cluttered scenes with robust, dense prehensile action poses (grasp poses and suction poses) and mask annotations. Of all the 100k point clouds, 80k point clouds as training data, 20k point clouds as testing dataset. Prehensile action branch (grasp and suction generation) are trained in the extended GPR dataset. The segmentation part of nonprehensile action branch (push generation) is trained in POIS \cite{xu2021pois} dataset. 

\textbf{The Extended GPR Dataset.} 
In grasp action of prehensile branch, the grasp quality metrics (grasp score) and grasp annotation are same with GPR \cite{wei2021gpr}, using Ferrari Canny metric \cite{ferrari1992planning} for labelling grasp quality.
In suction action of prehensile branch, 
we mainly choose seal formation score in dexnet 3.0 \cite{mahler2018dex}, which proposes a compliant suction contact model that computes the quality of the seal between the suction cup and local target surface, as the suction quality metrics (suction scores) to measure the quality of suction configuration.
A 6-DoF suction configuration is represented as suction center (contact point) $\bm L = (x, y, z)$ and suction direction (the normal direction of contact point) $\bm \Theta = \theta_{1}, \theta_{2}, \theta_{3}$.
Given a suction configuration for a single object, We first project the ring of the suction cup onto the surface of the object to get the contact ring with $n$ projecting contact points. Then, we use the $n$ projecting contact points to fit a plane $O$ by the least-squares method. We calculate the
Standard Deviation (SD) $\sigma$ of projection distance d by $\sigma=\sqrt{\frac{1}{n} \sum_{i=1}^{n}\left(d_{i}-\bar{d}\right)^{2}}$, where $\bar{d}$ denotes the mean
projection distance across all $n$ projecting contact points to plane $O$. 
The suction score $S_{s}$ can be calculate by the following formula:
\begin{equation}
S_{s}= e^{-(b \cdot \sigma)} \cdot \cos \left(\mathbf{M}, \mathbf{M}_{\mathbf{1}}\right)
\end{equation}
where b are constants for normalize, where $\mathbf{M}$ is the opposite direction of normal direction of plane $O$, $\mathbf{M}_{\mathbf{1}}$ is the suction direction of suction contact point. A suction annotation of single object can be organized as $L\left(p_i\right)=\left[\bm{L}, \boldsymbol{\bm \Theta}, S_s\right]$. 
\begin{equation}
\begin{gathered}
M\left(p_i\right)=\left[\mathbb{I}\left(f\left(S_{s}^{i}\right)\right)\right], \\
f\left(S_{s}^{i}\right)= \begin{cases}1 & \text { if } S_{s}^{i}>u, \\
0 & \text { otherwise }\end{cases}\label{point_label}
\end{gathered}
\end{equation}
Then, we adopt BlenderProc\cite{denninger2019blenderproc} to generate densely cluttered scenes of multi objects randomly. If no collision occurs in clutter and the suction score of point $i$ is greater than the threshold $u$, the contact point $i$ will be added into positive suction contact points set $M$, otherwise, the point will be added into negative suction contact points set, this process is organize as Eq.\ref{point_label}.

\section{Experiments}
In this section, we introduce the experimental setup details. Then, we 
carry out experiments to evaluate the effectiveness of PGS on dataset. Finally, we evaluate our PGS in Yumi IRB-1400 Robot platform, experimental results show that our model generalizes well to the real robot platform.

\subsection{Experimental Setup}

For prehensile action branch, 16384 points of point cloud cluttered scene are sampled as input. We use Adam optimizer with a mini-batch size of 8. The learning rate is set to 0.02 at start, and it is divided by 10 when the error plateaus. 
Nonprehensile action branch is trained for 30K iterations with Adam, with
an initial learning rate of 1e-4 and a batch size of 8.

\begin{figure}[bp]
    \vspace{-0.2in}
    \centering
    \subfigure[]{\includegraphics[width=0.41\linewidth]{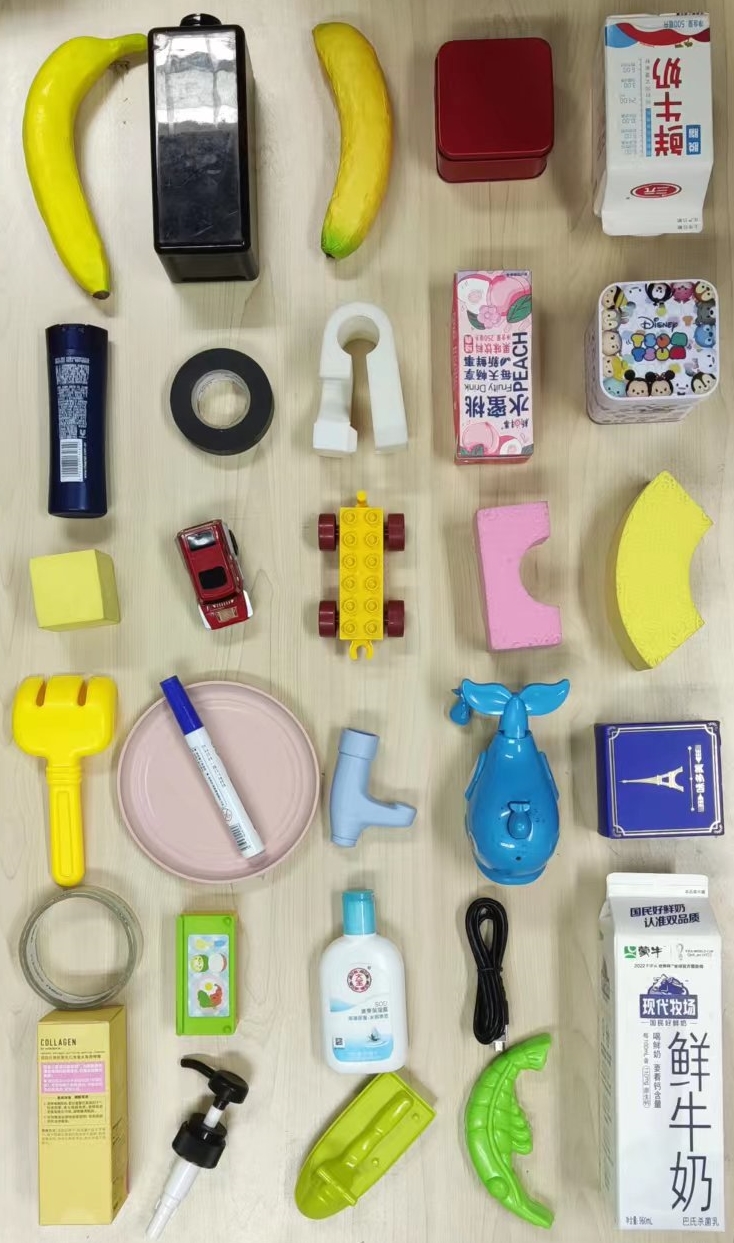}
    \label{fig:objects_and_robot_a}}
    \subfigure[]{\includegraphics[width=0.52\linewidth]{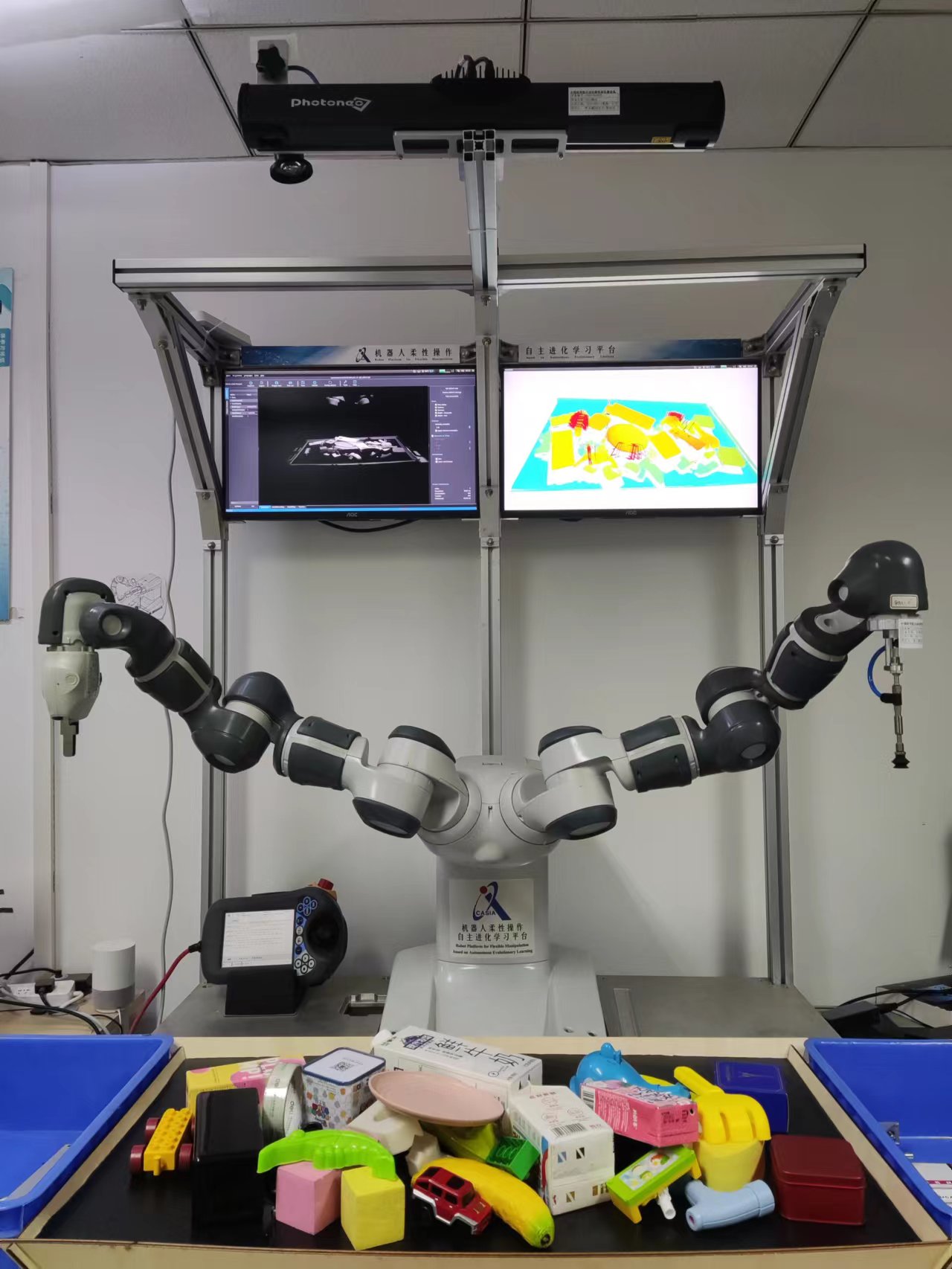}
    \label{fig:objects_and_robot_b}}
    \caption{Real setting of our robotic picking experiments. \subref{fig:objects_and_robot_a} Objects in our robotic experiments. \subref{fig:objects_and_robot_b} Cluttered scene setup on ABB Yumi robot system in picking experiment.}
    \label{fig:objects_and_robot}
    \vspace{-4mm}
\end{figure}

\subsection{Experimental Results}
In this section, we separately compare  grasp part, suction part and push's segmentation part of PGS with other state-of-the-art methods. 

\begin{table}[!t]
  \centering
  \caption{Grasp Comparation on Extended GPR Dataset}
  \vspace{+0.1in}
    \begin{tabular}{p{2.2cm}<{\centering}p{1.4cm}<{\centering}p{1.4cm}<{\centering}p{1.5cm}<{\centering}}
    \toprule
    Methods & \multicolumn{1}{c}{AP} & \multicolumn{1}{c}{AP$_{0.8}$} & \multicolumn{1}{c}{AP$_{0.4}$} \\
    \midrule
    GPD \cite{ten2017grasp}   & \multicolumn{1}{c}{26.6} & \multicolumn{1}{c}{31.5} & \multicolumn{1}{c}{16.7} \\
    PointNetGPD \cite{liang2019pointnetgpd} & \multicolumn{1}{c}{29.9} & \multicolumn{1}{c}{37.1} & \multicolumn{1}{c}{17.9} \\
    GraspNet\cite{fang2020graspnet} & \multicolumn{1}{c}{32.7} & \multicolumn{1}{c}{40.1} & \multicolumn{1}{c}{23.4} \\
    GPR\cite{wei2021gpr}   & \multicolumn{1}{c}{35.4} & \multicolumn{1}{c}{41.6} & \multicolumn{1}{c}{24.4} \\
    PGS   &  37.8 & 44.8 & 24.9  \\
    \bottomrule
    \end{tabular}%
    \vspace{-0.35cm}
  \label{tab:mainresults-grasp}%
\end{table}%

\begin{table}[!t]
  \centering
  \vspace{+0.05in}
  \caption{Suction Comparation on  Extended GPR Dataset}
  \vspace{+0.1in}
    \begin{tabular}{p{2.2cm}<{\centering}p{1.4cm}<{\centering}p{1.4cm}<{\centering}p{1.5cm}<{\centering}}
    \toprule
    Methods & \multicolumn{1}{c}{AP} & \multicolumn{1}{c}{AP$_{0.8}$} & \multicolumn{1}{c}{AP$_{0.4}$} \\
    \midrule
    Normal STD \cite{cao2021suctionnet} & \multicolumn{1}{c}{28.4} & \multicolumn{1}{c}{5.3} & \multicolumn{1}{c}{37.6} \\
    DexNet 3.0 \cite{mahler2018dex} & \multicolumn{1}{c}{26.6} & \multicolumn{1}{c}{4.8} & \multicolumn{1}{c}{35.6} \\
    DexNet 4.0 \cite{mahler2019learning} & \multicolumn{1}{c}{37.9} & \multicolumn{1}{c}{13.4} & \multicolumn{1}{c}{50.2} \\
    SuctionNet\cite{cao2021suctionnet} & \multicolumn{1}{c}{43.1} & \multicolumn{1}{c}{12.5} & \multicolumn{1}{c}{57.6} \\
    PGS   &   47.1    &  12.7     &  59.4 \\
    \bottomrule
    \end{tabular}%
    \vspace{-0.35cm}
  \label{tab:mainresults-suction}%
\end{table}%

\begin{table}[!t]
  \centering
  \vspace{+0.05in}
  \caption{Segmentation Comparation on POIS Dataset}
    \vspace{+0.1in}
    \begin{tabular}{p{3.2cm}<{\centering}p{1.1cm}<{\centering}p{1.1cm}<{\centering}p{1.1cm}<{\centering}}
    \toprule
    Methods & P     & R     & F \\
    \midrule
    Mask R-CNN (Depth) \cite{he2017mask} & 88.4  & 90.2  & 89.4 \\
    PGS   & 91.0    & 93.2  & 91.9 \\
    \bottomrule
    \end{tabular}%
    \vspace{-0.45cm}
  \label{tab:mainresults-segementation}%
\end{table}%

\begin{figure*}[!t]
	\centering
	\vspace{+0.4cm}
	\includegraphics[width=1.\textwidth]{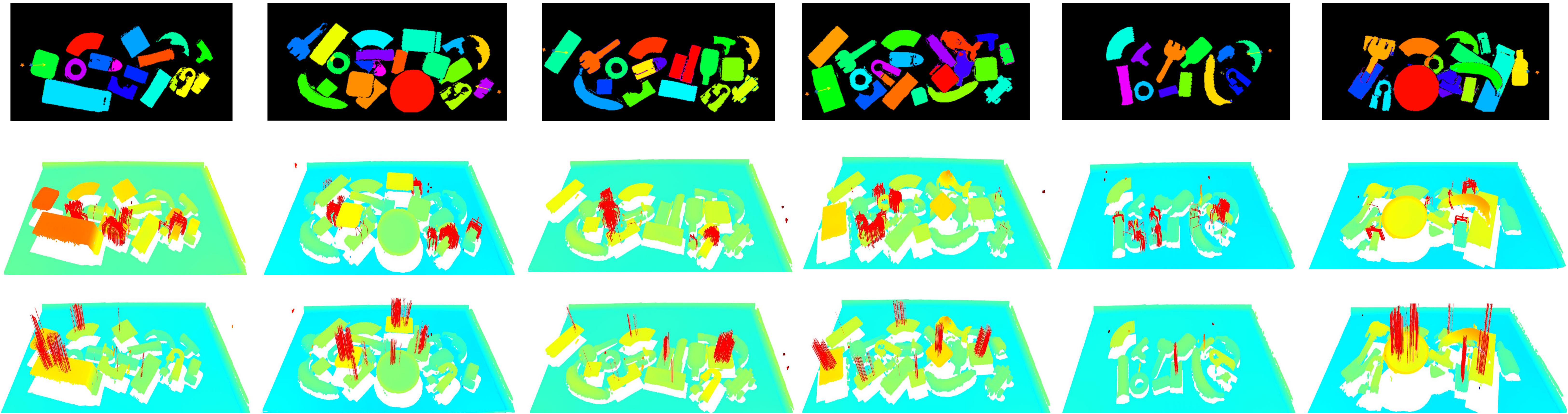}
	\vspace{+0.4cm}
	\caption{The results visualization of PGS in
real picking scene. The first line shows segmentation and push action results of scene depth image, the second and last line show grasp and suction generation of scene point cloud separately.}
	\label{results-visualization}
\end{figure*}

\begin{figure*}[!t]
	\centering
	\includegraphics[width=1.\textwidth]{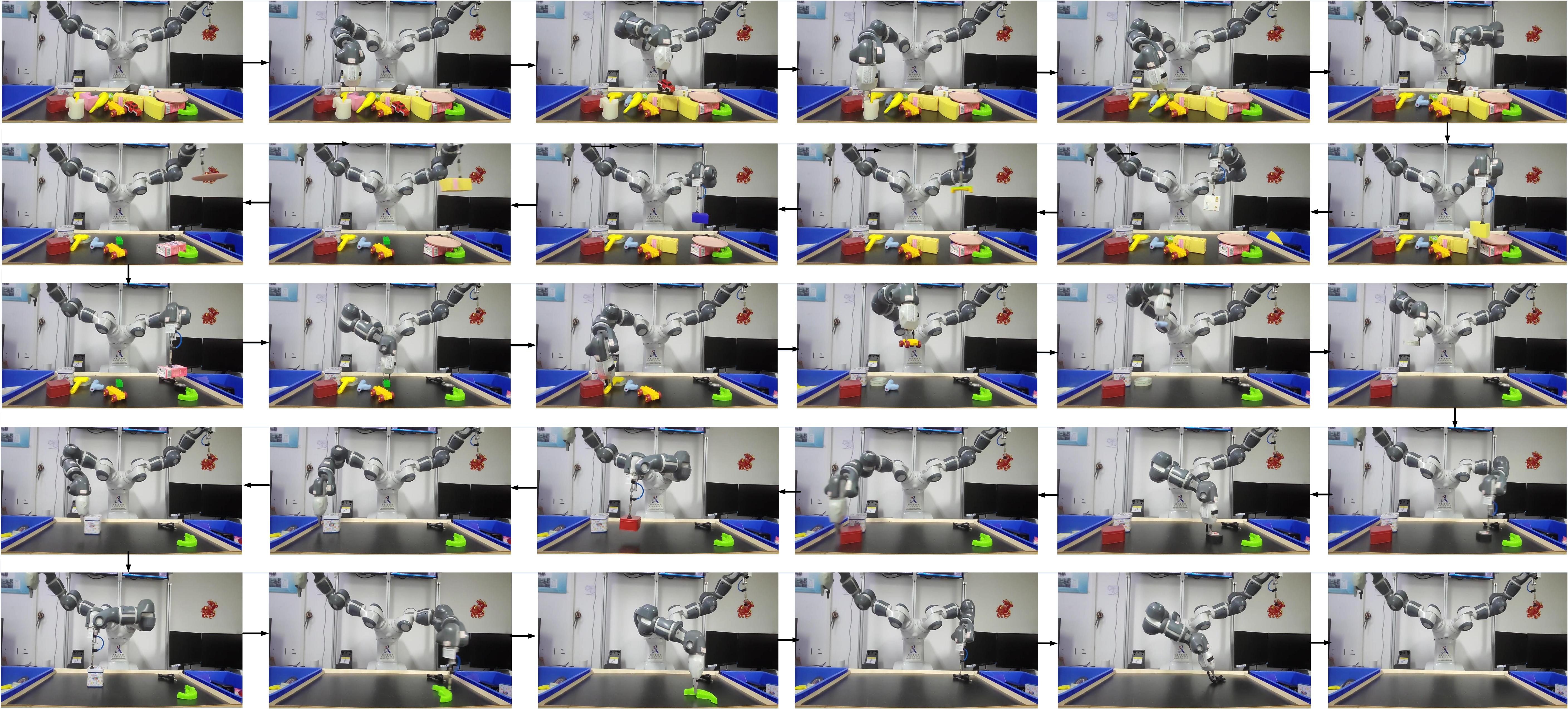}
	\vspace{+0.2cm}
	\caption{Qualitative experimental results of PGS in clutter picking. Robot system could carry out many ways (grasp, suction, push-to-grasp, push-to-suction) to clear objects in cluttered scene effectively and fluently. }
	\label{fig:Qualitative}
	\vspace{-0.6cm}
\end{figure*}

\textbf{Grasp Comparation.} The grasp comparation (only grasp) on extended GPR dataset is shown in Tab.\ref{tab:mainresults-grasp}. We compare with GPD \cite{ten2017grasp}, PointNetGPD \cite{liang2019pointnetgpd}, GraspNet\cite{fang2020graspnet} and GPR\cite{wei2021gpr}.
The evaluation metric for evaluate the quality of predicted grasps is according to GraspNet \cite{fang2020graspnet}.
Our method achieves the best performance in dataset which demonstrates the superiority of our method.

\textbf{Suction Comparation.} The suction comparation (only suction) on extended GPR dataset is shown in Tab.\ref{tab:mainresults-suction}. We compare with Normal STD \cite{cao2021suctionnet}, DexNet 3.0 \cite{mahler2018dex}, DexNet 4.0 \cite{mahler2019learning}, SuctionNet \cite{cao2021suctionnet}.
The evaluation metric for evaluate the quality of predicted suctions is according to SuctionNet \cite{cao2021suctionnet}.
The experiment shows our method also achieve the best performance in dataset.

\textbf{Segmentation Comparation.} The segmentation comparation on POIS dataset is shown in Tab.\ref{tab:mainresults-segementation}. We compare with Mask RCNN \cite{he2017mask} with depth input. We follow overlap P/R/F metrics in \cite{xie2021unseen} to evaluate the object segmentation performance. Result shows PGS achieve a great segmentation performance. 

\subsection{Robotic Experiments}

We validate the reliability and efficiency of our proposed
PGS network in ABB Yumi IRB-1400 with a parallel two-fingered gripper and a silicone suction cup. The max opening width of the gripper is equal to 50 mm. Objects are presented to the robot in dense clutter as shown in Fig. \ref{fig:objects_and_robot_a}. Robot system is shown in Fig. \ref{fig:objects_and_robot_b}. A high- resolution Photoneo PhoXi industrial sensor (1032x772 with 0.05 mm depth precision) is set 1300mm above the table.
Qualitative results shown in Fig.\ref{results-visualization} and Fig. \ref{fig:Qualitative}. Fig.\ref{results-visualization} are the results visualization of PGS in real picking scene, achieving good action generating results.
Fig. \ref{fig:Qualitative} are the picking process of PGS when clearing objects in real clutter scene, which
demonstrates our methods can carry out many ways (grasp, suction, push-to-grasp, push-to-suction) to complete clutter scenes objects picking effectively and fluently.
When objects out of  prehensile action work space, push action can be carried out to move object towards scene center, that is, PGS finish these objects picking through push-to-grasp or push-to-suction.

\begin{table}[tbp]
  \vspace{+0.1in}
  \centering
  \caption{Results of Robot Picking in Cluttered Real Scene.}
      \vspace{+0.1in}
    \begin{tabular}{p{2.2cm}<{\centering}p{0.6cm}<{\centering}p{0.6cm}<{\centering}p{0.6cm}<{\centering}p{1.0cm}<{\centering}p{1.0cm}<{\centering}}
    \toprule
    Methods & Grasp & Suction & Push &\multicolumn{1}{c}{SR ($\%$)} & \multicolumn{1}{c}{CR ($\%$)} \\
    \midrule
    GPD \cite{ten2017grasp}   & Yes& No& No & 48.8    &  55.7 \\
    PointNetGPD \cite{liang2019pointnetgpd} &Yes& No& No &  50.6    &  56.8 \\
    GraspNet\cite{fang2020graspnet}  &Yes& No& No  &  57.9    &  63.0 \\
    GPR\cite{wei2021gpr}         & Yes& No& No & 58.8    &  63.4 \\
    Normal STD \cite{cao2021suctionnet}  & No& Yes& No & 57.2    &  65.3 \\
    SuctionNet \cite{cao2021suctionnet} & No& Yes& No & 62.5    &  69.5 \\
    POIS \cite{xu2021pois}        & Yes& Yes& No & 72.5    &  83.9 \\
    DexNet 4.0 \cite{mahler2019learning}  &Yes& Yes& No &  82.3    &  86.4 \\
    PGS (ours)         & Yes& Yes& Yes & 89.6    &  96.2 \\
    \bottomrule
    \end{tabular}%
  \vspace{-0.55cm}
  \label{tab:Quantitative}%
\end{table}%
We compare PGS with state-of-the-art, open-sourced picking methods in real scene, including single prehensile action method (only grasp or only suction) like GPD \cite{ten2017grasp}, PointNetGPD \cite{liang2019pointnetgpd}, GraspNet\cite{fang2020graspnet} , GPR\cite{wei2021gpr}, Normal STD \cite{cao2021suctionnet}, SuctionNet \cite{cao2021suctionnet}, 
and two-type prehensile action (both grasp and suction) method 
POIS \cite{xu2021pois}, DexNet 4.0 \cite{mahler2019learning}. 
For each experiment procedure, we randomly choose (14, 17, 20) objects in Fig. \ref{fig:objects_and_robot_a}, and put them on a table to form a structure cluttered scene. The robot attempts multiple prehensile actions (grasp or suction) until all objects are picked or (19, 22, 25) prehensile actions have been attempted.
For the object out of prehensile action work space, two times push action are limited for one object picking, otherwise, it is recorded as a failure.
Actually in our experiment, push action mostly is executed one or two times for the object out of prehensile action work space.
We repeat the experiments 3 times for each method.
The statistical result is shown in Tab. \ref{tab:Quantitative}.
Success Rate (SR) and Completion Rate (CR) are used as the performance evaluation metrics, which represent the percentage of successful grasps and the percentage of objects removed from the clutter. 

The experiment shows our method PGS outperforms baseline methods in terms of success rate and completion rate by a large margin, which demonstrates the superiority of PGS.
Note that our experimental table is lager than the table of these comparing methods in original paper (about 2 times length size of single-arm methods, like GPR, and about 1.5 times length size of other dual-arm methods, like POIS). This is why the SR and CR of these comparing methods is lower than results in original paper. This is also the core meaning of adding push action into dual-arm robot picking, which could expand the picking space for ambidextrous robot. It is the main reason of our method outperforms than other two-type prehensile action method.

Our method outperforms than single prehensile action method by a large margin mainly contain two aspects: 1) single prehensile action cannot meet the requirement of multiple objects picking (like grasp action is  hard to pick large and flat object due to the wide limitation of grasp gripper, suction action is hard to pick small and hollowed-out objects due to air tightness requirement of suction cup). 2) without push action, object out of single-arm work space can not be picked by these method.

\section{Conclusion}
In this paper, we proposes a Push-Grasp-Suction (PGS) tri-mode grasping learning network for ambidextrous robot picking through the fusion of different prehensile actions and the fusion of prehensile action and nonprehensile action.
The prehensile branch of PGS takes point clouds as input, and the 6-DoF picking configuration of grasp and suction in cluttered scenes are generated by multi-task point cloud learning. The nonprehensile branch with depth image input generates instance segmentation map and push configuration, cooperating with the prehensile actions to complete the picking of objects out of single-arm space. PGS generalizes well in real scene and achieves state-of-the-art picking performance.
The experimental results verified the effectiveness of our method.

\bibliographystyle{IEEEtran}
\bibliography{IEEEabrv,reference}

\end{document}